\documentclass{article}

\usepackage{microtype}
\usepackage{graphicx}
\graphicspath{{./}{../}}
\usepackage{subcaption}
\usepackage{booktabs} 

\usepackage{hyperref}

\usepackage[accepted]{icml2026-workshop}

\usepackage{amsmath}
\usepackage{amssymb}
\usepackage{mathtools}
\usepackage{amsthm}
\usepackage{multirow}
\usepackage[normalem]{ulem}
\useunder{\uline}{\ul}{}
\usepackage{adjustbox}

\usepackage[capitalize,noabbrev]{cleveref}

\theoremstyle{plain}

\theoremstyle{definition}

\theoremstyle{remark}

\usepackage[textsize=tiny]{todonotes}

\icmltitlerunning{Semantics-Enhanced Retrieval-Augmented Time Series Forecasting}

\begin{document}

\icmlsetsymbol{siemensnote}{\dag}
\twocolumn[
  \icmltitle{Semantics-Enhanced Retrieval-Augmented Time Series Forecasting}

  \begin{icmlauthorlist}
    \icmlauthor{Shiqiao Zhou}{bhamcs}
    \icmlauthor{Zipeng Wu}{bhammath}
    \icmlauthor{Holger Schöner}{siemens}
    \icmlauthor{Edouard Fouché}{siemens,siemensnote}
    \icmlauthor{IAG Wilson}{bei}
    \icmlauthor{Shuo Wang}{bhamcs}
  \end{icmlauthorlist}

  \icmlaffiliation{bhamcs}{School of Computer Science, University of Birmingham, Birmingham, United Kingdom}
  \icmlaffiliation{bhammath}{School of Mathematics, University of Birmingham, Birmingham, United Kingdom}
  \icmlaffiliation{siemens}{Siemens AG, Munich, Germany}
  \icmlaffiliation{bei}{Birmingham Energy Institute, University of Birmingham, Birmingham, United Kingdom}

  \icmlcorrespondingauthor{Shuo Wang}{s.wang.2@bham.ac.uk}

  \icmlkeywords{Time Series Forecasting, Retrieval-Augmented Generation, Multimodal Retrieval}

  \vskip 0.3in
]

\printAffiliationsAndNotice{\textsuperscript{\dag}The contribution was carried out while this co-author was employed by Siemens AG.}

\begin{abstract}
Time series forecasting models often benefit from historical patterns. Inspired by Retrieval-Augmented Generation (RAG), recent research explored retrieving relevant historical time series segments to enhance forecasting. However, relying solely on time series similarity is often insufficient for retrieval under non-stationarity. To address this, we propose a multimodal approach: a \textbf{S}emantics-\textbf{E}nhanced \textbf{R}etrieval-\textbf{A}ugmented Time Series \textbf{F}orecasting framework, SERAF. Unlike mainstream approaches that depend only on time series similarity, SERAF conducts dual retrieval over the time series and their self-generated textual descriptions. It retrieves two complementary sets of historical patterns and corresponding futures, which are selectively and jointly used to guide future predictions. Experiments across seven real-world datasets demonstrate the effectiveness of SERAF in bridging numerical and semantic views of time series compared with state-of-the-art baselines.
\end{abstract}

\section{Introduction}
\label{sec:intro}
Multivariate time series forecasting predicts future trajectories from historical observations and is central to traffic~\citep{lippi2013short}, energy~\citep{daut2017building}, finance~\citep{poon2003forecasting}, and climate~\citep{price2025probabilistic}. Methods have progressed from classical statistical models such as ARIMA~\citep{box2015time} to deep forecasters such as DLinear, a robust MLP-based linear model~\citep{zeng2023transformers}, and PatchTST, which uses patch-level time series representations~\citep{nietime}. Recently, LLM-based methods such as Time-LLM~\citep{jin2023time} and GPT4TS~\citep{zhou2023one} have used textual context to inject background knowledge into forecasting.

The most widely adopted approach for supporting context-aware generation is retrieval-augmented generation (RAG), which retrieves relevant documents from large external databases and has become a key component of modern LLM pipelines~\citep{lewis2020retrieval}. Motivated by RAG, recent studies on time series forecasting have explored retrieval-based approaches that construct historical databases to retrieve similar patterns, thereby explicitly leveraging the entire history to guide future prediction. Representative efforts along this direction include RAFT, introducing multi-periodicity for retrieval-based forecasting~\citep{hanretrieval}; TimeRAG, integrating retrieved sequences into LLM-based forecasters~\citep{yang2025timerag}; TRACE, aligning external text with time series for multimodal retrieval~\citep{chen2025trace}; TS-RAG, augmenting time series foundation models (TSFMs) via adaptive retrieval~\citep{ning2025ts}; and TimeRAF, incorporating channel prompting into retrieval-augmented TSFMs~\citep{zhang2025timeraf}. 

However, for challenging non-stationary time series, most retrieval methods remain confined to time series 
similarity~\citep{hanretrieval,yang2025timerag, ning2025ts, zhang2025timeraf}, while multimodal approaches often rely on large external text corpora or LLM-generated descriptions, leading to inefficiency and limited scalability~\citep{chen2025trace}. To address this gap, we propose SERAF, a \textbf{S}emantics-\textbf{E}nhanced \textbf{R}etrieval-\textbf{A}ugmented Time Series \textbf{F}orecasting framework. Beyond retrieving relevant patterns via TS similarity, SERAF introduces a semantic retrieval module based on textual descriptions generated directly from time series segments. SERAF performs retrieval from both temporal and semantic perspectives, and adaptively fuses the retrieved results to enhance forecasting, all within a lightweight pipeline that requires no external annotations or domain-specific texts. 

This is useful because two time series segments may differ in raw scale or local shape while sharing high-level attributes such as season, trend, and volatility. By indexing these attributes as text, SERAF can retrieve semantically relevant historical futures that are not necessarily nearest under TS similarity, complementing numerical retrieval without relying on external text resources.

Our main contributions are summarized as follows:
\begin{itemize}
\item We propose a novel semantic retrieval strategy based on textual descriptions, enriching the retrieval process beyond purely numerical time series similarity.
\item The textual descriptions are automatically generated from time series segments, requiring no external texts, thus ensuring efficiency and scalability.
\item Extensive experiments on seven real-world datasets demonstrate SERAF's forecasting improvements.
\end{itemize}

\begin{figure*}[h]
  \centering
  \includegraphics[width=\linewidth, trim = 0cm 3cm 0cm 0.4cm, clip]{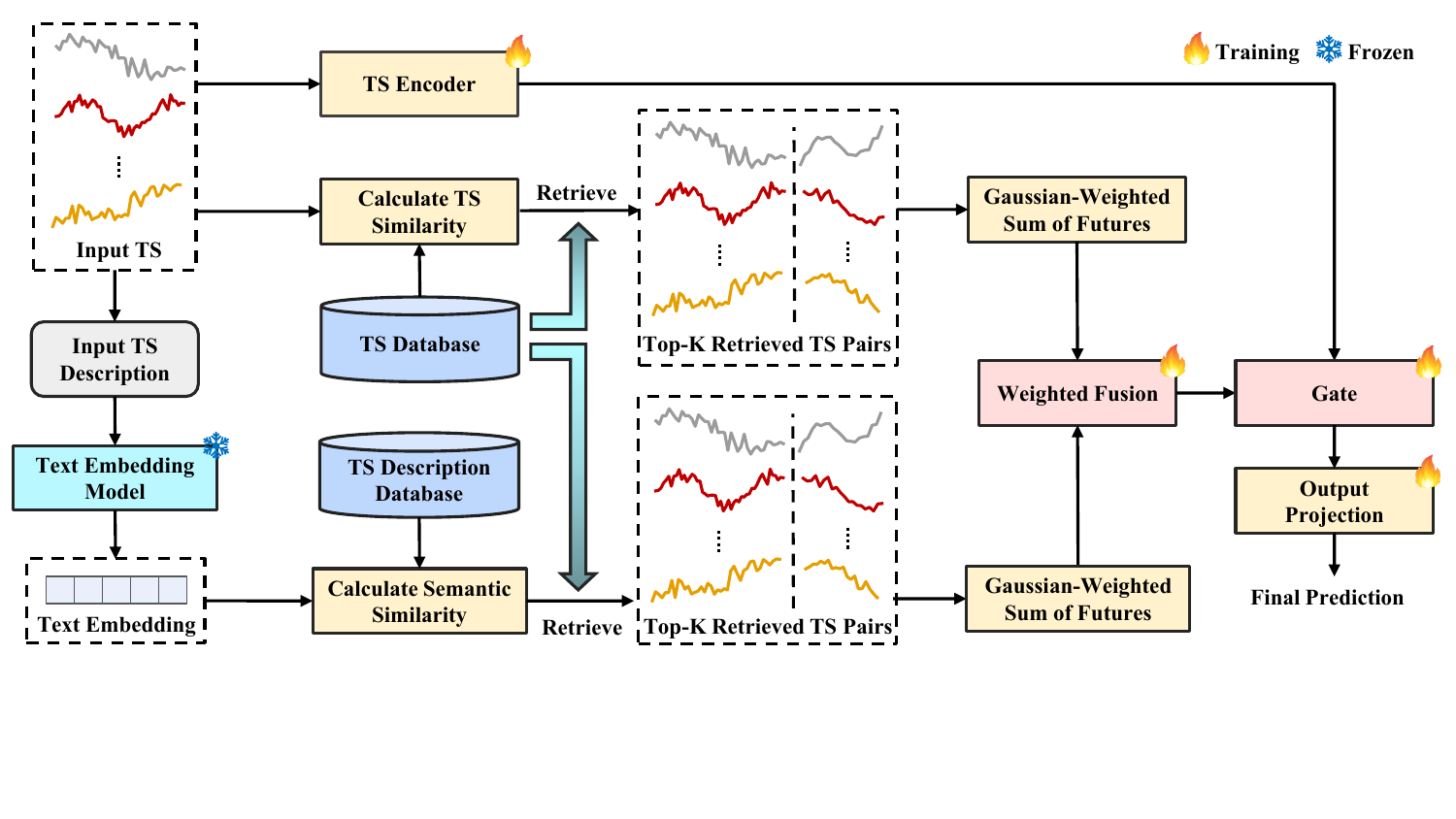}
  \caption{Overview of SERAF. Each retrieved TS pair contains a historical time series segment and its corresponding future.}
  \label{SERAF}
\end{figure*}

\begin{figure}[h]
  \centering
  \includegraphics[width=0.8\linewidth, trim = 8cm 6.8cm 9.6cm 4.5cm, clip]{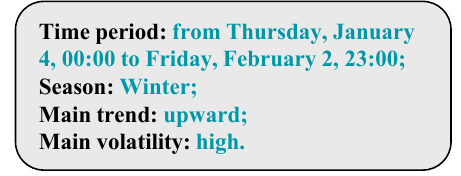}
  \caption{An example of textual description of time series.}
  \label{fig_text}
\end{figure}

\section{Method}
\label{sec:method}
\subsection{Problem Formulation}

Given a historical input sequence $\mathbf{X}_{t-L+1:t} = \{X_{t-L+1}, X_{t-L+2}, \dots, X_t\} \in \mathbb{R}^{L\times C}$, where $L$ is the look-back window and $C$ is the number of channels, the goal is to predict $\mathbf{Y}_{t+1:t+H} = \{Y_{t+1}, Y_{t+2}, \dots,\ Y_{t+H}\}$ over the next $H$ time steps.


\subsection{Overview}
As shown in Figure~\ref{SERAF}, SERAF performs retrieval-augmented forecasting from both temporal and semantic information. Given an input time series, a trainable encoder produces a naive prediction, while TS-similar historical segments are retrieved from a time series database. In parallel, a textual description of the input is embedded by a frozen text model to retrieve semantically similar descriptions from an aligned description database. The retrieved futures from both modalities are Gaussian-weighted, fused, gated with the naive prediction, and projected to produce the final prediction. This semantic retrieval dimension enriches the search space beyond temporal patterns and improves forecasting.

\subsection{TS Database and TS Description Database}
We construct the Time Series (TS) database by sliding a length-$L$ window over the training set with stride 1 for dense historical coverage. Each segment is paired with its future sequence of length $H$. The TS database is $D_T = \{(\mathbf{P}_T^i, \mathbf{F}_T^i)\}_{i=1}^N$, where each pair contains a historical segment $\mathbf{P}_T^i$ and its future $\mathbf{F}_T^i$.

In parallel, we build a TS Description Database by generating a natural language description for each pair in $D_T$ with a predefined template. As illustrated in Figure~\ref{fig_text}, each description includes time period, season, main trend, and main volatility. Time period and season come from timestamps, while trend and volatility use the most frequent channel-level pattern. Trend is categorized as upward, downward, or stable, and volatility as high, medium, or low. We denote the database as $D_S = \{\mathbf{Q}_S^i\}_{i=1}^N$, where each description $Q_S^i$ aligns with $\mathbf{P}_T^i$ in $D_T$.

\subsection{Retrieval from Time Series Similarity}
For TS retrieval, given the $j$-th input sequence $\mathbf{X}^j$, we compute the similarity score $\rho_{ij}$ between $\mathbf{X}^j$ and each historical segment $\mathbf{P}_T^i$ in $D_T$ using a similarity function $sim$: 
\begin{align}
\rho_{ij} = sim(\mathbf{X}^j, \mathbf{P}_T^i), \quad i \in [1,N].
\end{align}
We use Pearson's correlation because it reduces the effects of scale variation and value shifts while emphasizing monotonic trends~\citep{hanretrieval}. 
To avoid leakage, historical segments in $D_T$ that overlap with the input are excluded during training. The valid index set is:
\begin{equation}
\mathcal{I}_{\text{valid}} =
\left\{
\begin{array}{@{}l@{\;}l}
\{&i \in [1,N] \mid i \notin [\max(1, j-(L+H-1)), \\&\min(N, j+(L+H-1))] \,\}, \, \text{if training}, \\[6pt]
\{&i \in [1,N] \,\}, \,  \text{otherwise}.
\end{array}
\right.
\label{searching}
\end{equation}
The Top-$K$ index set $\mathcal{K}_T^j$ and retrieval set $\mathcal{R}_T^j$ are:
\begin{align}
\mathcal{K}_T^j
   = \text{Top-}K \{ \rho_{ij} \mid i \in \mathcal{I}_{\text{valid}} \}, \ \ |\mathcal{K}_T^j| = K, \
\label{topk}
\end{align}
\begin{equation}
\mathcal{R}_T^j = \{\, ( \mathbf{P}_T^{k}, \mathbf{F}_T^{k} ) \mid 
   k \in \mathcal{K}_T^j  \,\}.
\label{retrieval set}
\end{equation}
The $K$ retrieved segments are weighted by a Gaussian kernel, assigning larger weights to higher similarities:
\begin{equation}
\alpha_{T}^{kj} = 
\frac{\exp\!\left(-\tfrac{(1-\rho_{kj})^2}{2\tau^2}\right)}
{\sum\limits_{i \in \mathcal{K}^j} \exp\!\left(-\tfrac{(1-\rho_{ij})^2}{2\tau^2}\right)},
\label{weight}
\end{equation}
where $\tau$ is the Gaussian bandwidth. 

The TS-similarity retrieved future is:
\begin{equation}
\hat{\mathbf{F}}_T^j \;=\; 
\sum_{k \in \mathcal{K}_T^j} \alpha_{T}^{kj} \, \mathbf{F}_T^{k}.
\label{weighted sum}
\end{equation}

\subsection{Retrieval from Semantic Similarity}
For semantic retrieval, we generate a textual description $\mathbf{Q}^j$ for input $\mathbf{X}^j$ using the same template as in Figure~\ref{fig_text}. This extracts attributes from the time series itself without external text. A frozen text embedding model encodes $\mathbf{Q}^j$ into $\mathbf{E}^j$ and each historical description $\mathbf{Q}_S^i$ into $\mathbf{E}_S^i$. Semantic similarity is computed as:
\begin{align}
s_{ij} = sim(\mathbf{E}^j, \mathbf{E}_S^i), \quad i \in [1,N],
\end{align}
where $sim$ is cosine similarity. Analogous to Equation~\ref{retrieval set}, we retrieve Top-$K$ TS pairs and define:
\begin{equation} 
\mathcal{R}_S^j = \{\, ( \mathbf{P}_T^{k}, \mathbf{F}_T^{k} ) \mid k \in \mathcal{K}_S^j \,\}, 
\label{retrieval set_2} 
\end{equation}
where $\mathcal{K}_S^j$ is the Top-$K$ index set. The retrieved segments are weighted as in Eq.~\ref{weight}, giving:
\begin{equation} 
\hat{\mathbf{F}}_S^j \;=\; \sum_{k \in \mathcal{K}_S^j} \alpha_{S}^{kj} \, \mathbf{F}_T^{k}. 
\label{weighted sum_2} 
\end{equation}

\subsection{Fusion and Final Prediction}
To adaptively balance the contributions of semantic and temporal retrieval, their aggregated futures $\hat{\mathbf{F}}_S^j$ and $\hat{\mathbf{F}}_T^j$ are first fused with a learnable weight:
\begin{equation}
\hat{\mathbf{F}}^j = w \, \hat{\mathbf{F}}_S^j + (1-w) \, \hat{\mathbf{F}}_T^j,
\end{equation}
where $w\in (0,1)$ is a trainable parameter.

In addition to the two retrieval modules, the input $X^j$ is passed through a linear time series (TS) encoder to produce a naive prediction $\hat{\mathbf{X}}^j$. This output is then adaptively integrated with $\hat{\mathbf{F}}^j$ through a gating mechanism, which dynamically balances their relative contributions:
\begin{equation}
\mathbf{G}^j = \beta \ \hat{\mathbf{X}}^j + (1-\beta  )\hat{\mathbf{F}}^j,
\quad
\beta  = \sigma\left( W [\hat{\mathbf{X}}^j ; \hat{\mathbf{F}}^j] \right).
\end{equation}
where $[\,\cdot;\cdot\,]$ denotes concatenation, $W$ is a learnable projection, and $\sigma(\cdot)$ is the sigmoid function. 

Finally, the representation $\mathbf{G}^j$ is mapped by an output projection to the final prediction $\hat{\mathbf{Y}}^j$, and the model is trained by minimizing the mean squared error (MSE) loss.

\section{Experiment}
\subsection{Experimental Settings}
\textbf{Datasets}. 
We train and evaluate SERAF 
on seven widely used multivariate time series datasets: ETTh1, ETTh2, ETTm1, ETTm2, Exchange, Weather, and Electricity~\citep{wu2021autoformer}.

\begin{table*}[t]
\centering
\caption{Comparison of SERAF and baselines over seven datasets. All the results are with the same input time series length $=720$ 
and averaged across four horizons (96, 192, 336, 720). 
Best results are shown in \textbf{bold} and second-best results are \underline{underlined}.
}
\label{main_results}
\scriptsize
\setlength{\tabcolsep}{2.5pt}
\resizebox{\textwidth}{!}{%
\begin{tabular}{@{}l*{16}{c}@{}}
\toprule
& \multicolumn{2}{c}{SERAF} & \multicolumn{2}{c}{RAFT} & \multicolumn{2}{c}{CycleNet} & \multicolumn{2}{c}{PatchTST} & \multicolumn{2}{c}{DLinear} & \multicolumn{2}{c}{TimeMixer} & \multicolumn{2}{c}{TimesNet} & \multicolumn{2}{c}{Autoformer} \\
\cmidrule(lr){2-3}\cmidrule(lr){4-5}\cmidrule(lr){6-7}\cmidrule(lr){8-9}\cmidrule(lr){10-11}\cmidrule(lr){12-13}\cmidrule(lr){14-15}\cmidrule(l){16-17}
Dataset & MSE & MAE & MSE & MAE & MSE & MAE & MSE & MAE & MSE & MAE & MSE & MAE & MSE & MAE & MSE & MAE \\
\midrule
ETTh1 & \textbf{0.417} & \textbf{0.432} & \underline{0.418} & \underline{0.434} & 0.437 & 0.450 & 0.696 & 0.595 & 0.521 & 0.508 & 0.447 & 0.440 & 0.909 & 0.679 & 1.090 & 0.785 \\
ETTh2 & \textbf{0.348} & \underline{0.401} & \underline{0.358} & 0.408 & 0.368 & 0.406 & 0.472 & 0.479 & 0.445 & 0.472 & 0.365 & \textbf{0.395} & 0.486 & 0.491 & 0.494 & 0.516 \\
ETTm1 & \textbf{0.346} & \textbf{0.377} & \underline{0.347} & \textbf{0.377} & 0.365 & \underline{0.390} & 0.551 & 0.489 & 0.400 & 0.422 & 0.381 & 0.396 & 0.926 & 0.653 & 4.233 & 1.653 \\
ETTm2 & \textbf{0.252} & \textbf{0.317} & \underline{0.257} & \underline{0.321} & 0.285 & 0.338 & 0.354 & 0.383 & 0.290 & 0.346 & 0.275 & 0.323 & 0.418 & 0.426 & 1.363 & 0.830 \\
Weather & 0.235 & 0.280 & 0.241 & 0.287 & \textbf{0.224} & \textbf{0.266} & \underline{0.232} & 0.274 & 0.238 & 0.289 & 0.240 & \underline{0.272} & 0.475 & 0.421 & 0.424 & 0.449 \\
Exchange & \underline{0.419} & \underline{0.430} & 0.449 & 0.444 & \textbf{0.403} & \textbf{0.426} & 0.436 & 0.455 & 0.465 & 0.463 & 0.504 & 0.468 & 0.681 & 0.534 & 2.053 & 1.064 \\
Electricity & \textbf{0.156} & \underline{0.257} & 0.160 & 0.259 & \underline{0.157} & \textbf{0.252} & 0.216 & 0.318 & 0.225 & 0.319 & 0.169 & 0.268 & 0.193 & 0.304 & 0.227 & 0.364 \\
\bottomrule
\end{tabular}%
}
\end{table*}

\noindent\textbf{Baselines}. We compare SERAF against seven state-of-the-art time series forecasting models. Autoformer~\citep{wu2021autoformer} and PatchTST~\citep{nietime} are Transformer-based forecasters. TimeMixer~\citep{wang2024timemixer} is an MLP-based multiscale mixing model. DLinear~\citep{zeng2023transformers} employs a lightweight yet robust linear model. RAFT~\citep{hanretrieval} enhances linear models with multi-period retrieval modules based on TS similarity. CycleNet~\citep{lin2024cyclenet} explicitly captures periodic patterns on linear backbones. TimesNet~\citep{wutimesnet} detects dominant periods via Fourier analysis.

\noindent\textbf{Implementation details.}
We follow the experimental settings of RAFT~\citep{hanretrieval}. The batch size is set to 32, and the Adam optimizer is employed. The input length is set as 720 and $\tau$ is set as 0.1. We use all-MiniLM-L6-v2 as the text embedding model. All experiments are conducted on one NVIDIA A100 GPU. Each result reported in the tables is reproduced and averaged over three independent runs.

\subsection{Main Results}
The forecasting results are shown in Table~\ref{main_results}. 
SERAF achieves the best MSE on all four ETT datasets and the best or tied-best MAE on three of them, with second-best MAE on ETTh2. Compared with the retrieval-based baseline RAFT, SERAF matches or improves both metrics across all seven datasets, reducing the averaged MSE and MAE by $2.56\%$ and $1.42\%$, respectively. Compared with CycleNet, SERAF reduces the averaged MSE and MAE by $2.95\%$ and $1.34\%$, respectively. These results suggest that semantic retrieval consistently improves over TS-similarity retrieval, while its gains may be less pronounced on datasets whose future dynamics are harder to capture through coarse semantic descriptions alone. 


\subsection{Ablation Study}
We conduct three ablation studies on ETTh2 and ETTm2 datasets to evaluate the contributions of key components in SERAF, as shown in Table~\ref{ablation}. The results are averaged across four forecasting horizons (96, 192, 336, 720), with Full (SERAF) using the averaged results from Table~\ref{main_results}. Removing the semantic retrieval module (w/o text) leads to the most significant average drop, confirming the complementary benefits of semantic similarity beyond pure time series similarity. Excluding the gating mechanism (w/o gate) and replacing it with simple concatenation, as in RAFT~\citep{hanretrieval}, results in a general performance drop, highlighting the role of gating in adaptively balancing the encoder’s prediction with retrieval-enhanced signals. Finally, replacing the learnable weighted fusion with uniform averaging (w/o weight) also degrades performance, demonstrating the effectiveness of relevance-aware weighting in fusing the two retrieval results.
\begin{table}[t]
\centering
\caption{
Ablation study of SERAF on ETTh2 and ETTm2 datasets, with MSE and MAE averaged over four forecasting horizons (96, 192, 336, 720). Best results are shown in \textbf{bold}.
}
\label{ablation}
\scriptsize
\setlength{\tabcolsep}{3pt}
\resizebox{\columnwidth}{!}{%
\begin{tabular}{@{}l*{8}{c}@{}}
\toprule
& \multicolumn{2}{c}{Full (SERAF)} & \multicolumn{2}{c}{w/o text} & \multicolumn{2}{c}{w/o gate} & \multicolumn{2}{c}{w/o weight} \\
\cmidrule(lr){2-3}\cmidrule(lr){4-5}\cmidrule(lr){6-7}\cmidrule(l){8-9}
Dataset & MSE & MAE & MSE & MAE & MSE & MAE & MSE & MAE \\
\midrule
ETTh2 & \textbf{0.348} & \textbf{0.401} & 0.354 & 0.405 & 0.351 & 0.404 & 0.350 & 0.403 \\
ETTm2 & \textbf{0.252} & \textbf{0.317} & 0.254 & 0.319 & 0.253 & 0.318 & 0.252 & 0.318 \\
\bottomrule
\end{tabular}
}
\end{table}

\subsection{Inference Efficiency Analysis}
We profile inference-stage efficiency on ETTh1 with forecasting horizon 96 and batch size 32. All compared backbones are lightweight MLP-based forecasters. As shown in Table~\ref{efficiency}, SERAF uses the smallest forecasting backbone, with 0.088M parameters and 0.34 MiB model GPU memory, compared with 0.138M parameters and 0.53 MiB for DLinear and 0.097M parameters and 0.37 MiB for RAFT. SERAF achieves the fastest inference speed, requiring 0.0075 seconds per iteration, compared with 0.0203 and 0.0230 seconds per iteration for DLinear and RAFT, respectively.

\begin{table}[t]
\centering
\caption{Inference-stage efficiency profiling on ETTh1 with horizon 96 and batch size 32.}
\label{efficiency}
\scriptsize
\setlength{\tabcolsep}{3pt}
\resizebox{\columnwidth}{!}{%
\begin{tabular}{@{}lccc@{}}
\toprule
Method & Parameters (M) & Model GPU Memory (MiB) & Inference Time (s/iter) \\
\midrule
DLinear & 0.138 & 0.53 & 0.0203 \\
RAFT & 0.097 & 0.37 & 0.0230 \\
SERAF & \textbf{0.088} & \textbf{0.34} & \textbf{0.0075} \\
\bottomrule
\end{tabular}
}
\end{table}

\section{Conclusion}
In this paper, we proposed SERAF, which enhances retrieval-based forecasting by incorporating semantic retrieval with self-generated textual descriptions of time series. By constructing parallel time series and text databases, SERAF retrieves complementary historical patterns and adaptively fuses them through weighted fusion, while a gating module balances retrieval signals with the TS encoder’s initial prediction. Experiments on seven real-world datasets and ablation studies highlight SERAF’s design and demonstrate accuracy improvements. For future work, we will explore richer text templates and refined Top-$K$ retrieval to reduce redundancy and improve robustness, toward more interpretable and generalizable retrieval-augmented time series forecasting.


\bibliographystyle{icml2026}
\bibliography{refs}

@inproceedings{hanretrieval,
  title={Retrieval Augmented Time Series Forecasting},
  author={Han, Sungwon and Lee, Seungeon and Cha, Meeyoung and Arik, Sercan O and Yoon, Jinsung},
  booktitle={Forty-second International Conference on Machine Learning},
  year={2025}
}

@article{zeng2023transformers,
  title={Are Transformers Effective for Time Series Forecasting?},
  author={Zeng, Ailing and Chen, Muxi and Zhang, Lei and Xu, Qiang},
  journal={arXiv preprint arXiv:2205.13504},
  year={2022}
}

@inproceedings{nietime,
  title={A Time Series is Worth 64 Words: Long-term Forecasting with Transformers},
  author={Nie, Yuqi and Nguyen, Nam H and Sinthong, Phanwadee and Kalagnanam, Jayant},
  booktitle={The Eleventh International Conference on Learning Representations},
  year={2025}
}

@article{jin2023time,
  title={Time-llm: Time series forecasting by reprogramming large language models},
  author={Jin, Ming and Wang, Shiyu and Ma, Lintao and Chu, Zhixuan and Zhang, James Y and Shi, Xiaoming and Chen, Pin-Yu and Liang, Yuxuan and Li, Yuan-Fang and Pan, Shirui and others},
  journal={arXiv preprint arXiv:2310.01728},
  year={2023}
}

@article{zhou2023one,
  title={One fits all: Power general time series analysis by pretrained lm},
  author={Zhou, Tian and Niu, Peisong and Sun, Liang and Jin, Rong and others},
  journal={Advances in neural information processing systems},
  volume={36},
  pages={43322--43355},
  year={2023}
}

@article{daut2017building,
  title={Building electrical energy consumption forecasting analysis using conventional and artificial intelligence methods: A review},
  author={Daut, Mohammad Azhar Mat and Hassan, Mohammad Yusri and Abdullah, Hayati and Rahman, Hasimah Abdul and Abdullah, Md Pauzi and Hussin, Faridah},
  journal={Renewable and Sustainable Energy Reviews},
  volume={70},
  pages={1108--1118},
  year={2017},
  publisher={Elsevier}
}

@article{price2025probabilistic,
  title={Probabilistic weather forecasting with machine learning},
  author={Price, Ilan and Sanchez-Gonzalez, Alvaro and Alet, Ferran and Andersson, Tom R and El-Kadi, Andrew and Masters, Dominic and Ewalds, Timo and Stott, Jacklynn and Mohamed, Shakir and Battaglia, Peter and others},
  journal={Nature},
  volume={637},
  number={8044},
  pages={84--90},
  year={2025},
  publisher={Nature Publishing Group UK London}
}

@article{lippi2013short,
  title={Short-term traffic flow forecasting: An experimental comparison of time-series analysis and supervised learning},
  author={Lippi, Marco and Bertini, Matteo and Frasconi, Paolo},
  journal={IEEE Transactions on Intelligent Transportation Systems},
  volume={14},
  number={2},
  pages={871--882},
  year={2013},
  publisher={IEEE}
}

@book{box2015time,
  title={Time series analysis: forecasting and control},
  author={Box, George EP and Jenkins, Gwilym M and Reinsel, Gregory C and Ljung, Greta M},
  year={2015},
  publisher={John Wiley \& Sons}
}

@article{lewis2020retrieval,
  title={Retrieval-augmented generation for knowledge-intensive nlp tasks},
  author={Lewis, Patrick and Perez, Ethan and Piktus, Aleksandra and Petroni, Fabio and Karpukhin, Vladimir and Goyal, Naman and K{\"u}ttler, Heinrich and Lewis, Mike and Yih, Wen-tau and Rockt{\"a}schel, Tim and others},
  journal={Advances in neural information processing systems},
  volume={33},
  pages={9459--9474},
  year={2020}
}

@article{zhang2025timeraf,
  title={Timeraf: Retrieval-augmented foundation model for zero-shot time series forecasting},
  author={Zhang, Huanyu and Xu, Chang and Zhang, Yi-Fan and Zhang, Zhang and Wang, Liang and Bian, Jiang},
  journal={IEEE Transactions on Knowledge and Data Engineering},
  year={2025},
  publisher={IEEE}
}

@article{ning2025ts,
  title={Ts-rag: Retrieval-augmented generation based time series foundation models are stronger zero-shot forecaster},
  author={Ning, Kanghui and Pan, Zijie and Liu, Yu and Jiang, Yushan and Zhang, James Y and Rasul, Kashif and Schneider, Anderson and Ma, Lintao and Nevmyvaka, Yuriy and Song, Dongjin},
  journal={arXiv preprint arXiv:2503.07649},
  year={2025}
}

@article{chen2025trace,
  title={TRACE: Grounding Time Series in Context for Multimodal Embedding and Retrieval},
  author={Chen, Jialin and Zhao, Ziyu and Nurbek, Gaukhar and Feng, Aosong and Maatouk, Ali and Tassiulas, Leandros and Gao, Yifeng and Ying, Rex},
  journal={arXiv preprint arXiv:2506.09114},
  year={2025}
}

@inproceedings{yang2025timerag,
  title={Timerag: Boosting llm time series forecasting via retrieval-augmented generation},
  author={Yang, Silin and Wang, Dong and Zheng, Haoqi and Jin, Ruochun},
  booktitle={ICASSP 2025-2025 IEEE International Conference on Acoustics, Speech and Signal Processing (ICASSP)},
  pages={1--5},
  year={2025},
  organization={IEEE}
}

@article{wu2021autoformer,
  title={Autoformer: Decomposition transformers with auto-correlation for long-term series forecasting},
  author={Wu, Haixu and Xu, Jiehui and Wang, Jianmin and Long, Mingsheng},
  journal={Advances in neural information processing systems},
  volume={34},
  pages={22419--22430},
  year={2021}
}

@inproceedings{wutimesnet,
  title={TimesNet: Temporal 2D-Variation Modeling for General Time Series Analysis},
  author={Wu, Haixu and Hu, Tengge and Liu, Yong and Zhou, Hang and Wang, Jianmin and Long, Mingsheng},
  booktitle={The Eleventh International Conference on Learning Representations},
  year={2023}
}

@article{lin2024cyclenet,
  title={Cyclenet: Enhancing time series forecasting through modeling periodic patterns},
  author={Lin, Shengsheng and Lin, Weiwei and Hu, Xinyi and Wu, Wentai and Mo, Ruichao and Zhong, Haocheng},
  journal={Advances in Neural Information Processing Systems},
  volume={37},
  pages={106315--106345},
  year={2024}
}

@article{poon2003forecasting,
  title={Forecasting volatility in financial markets: A review},
  author={Poon, Ser-Huang and Granger, Clive W J},
  journal={Journal of economic literature},
  volume={41},
  number={2},
  pages={478--539},
  year={2003},
  publisher={American Economic Association}
}

@inproceedings{wang2024timemixer,
  title={TimeMixer: Decomposable Multiscale Mixing for Time Series Forecasting},
  author={Wang, Shiyu and Wu, Haixu and Shi, Xiaoming and Hu, Tengge and Luo, Huakun and Ma, Lintao and Zhang, James Y and Zhou, Jun},
  booktitle={The Twelfth International Conference on Learning Representations},
  year={2024}
}

\end{document}